\documentclass[journal]{IEEEtran}

\ifCLASSOPTIONcompsoc
  \usepackage[nocompress]{cite}
\else
  \usepackage{cite}
\fi

\usepackage{amsmath}
\DeclareMathOperator*{\argmax}{argmax}
\usepackage{algorithm}
\usepackage{algorithmic}
\usepackage{graphicx}
\usepackage{comment}
\usepackage{url}

\ifCLASSINFOpdf

\else
  
\fi

\begin{document}

\title{Distributive Dynamic Spectrum Access through Deep Reinforcement Learning: A Reservoir Computing Based Approach}

\author{Hao-Hsuan~Chang, Hao~Song, Yang Yi, Jianzhong (Charlie) Zhang, Haibo He, and Lingjia~Liu
\thanks{H. Chang, H. Song, Y. Yi and L. Liu are with the Bradley Department of Electrical and Computer Engineering, Virginia Tech, Blacksburg, VA 24060, USA. J. Zhang is with Samsung Research America, Richardson, TX 75081, USA. H. He is with the Department of Electrical, Computer, and Biomedical Engineering, University of Rhode Island, Kingston, RI 02881, USA.}
\thanks{The corresponding author is L. Liu (ljliu@ieee.org).}
}

\markboth{IEEE INTERNET OF THINGS JOURNAL}%
{Shell \MakeLowercase{\textit{et al.}}: Bare Demo of IEEEtran.cls for IEEE Journals}

\maketitle

\IEEEpubid{This article is accepted to IEEE IoT Journal 2018, but has not been fully edited. Content may change prior to final publication. Copyright (c) 2018 IEEE.
}

\IEEEpubidadjcol

\begin{abstract}
Dynamic spectrum access (DSA) is regarded as an effective and efficient technology to share radio spectrum among different networks.
As a secondary user (SU), a DSA device will face two critical problems: avoiding causing harmful interference to primary users (PUs), and conducting effective interference coordination with other secondary users.
These two problems become even more challenging for a distributed DSA network where there is no centralized controllers for SUs.
%
In this paper, we investigate communication strategies of a distributive DSA network under the presence of spectrum sensing errors.
%
To be specific, we apply the powerful machine learning tool, deep reinforcement learning (DRL), for SUs to learn ``appropriate'' spectrum access strategies in a distributed fashion assuming NO knowledge of the underlying system statistics.
Furthermore, a special type of recurrent neural network (RNN), called the reservoir computing (RC), is utilized to realize DRL by taking advantage of the underlying temporal correlation of the DSA network.
Using the introduced machine learning-based strategy, SUs could make spectrum access decisions distributedly relying only on their own current and past spectrum sensing outcomes.
Through extensive experiments, our results suggest that the RC-based spectrum access strategy can help the SU to significantly reduce the chances of collision with PUs and other SUs.
We also show that our scheme outperforms the myopic method which assumes the knowledge of system statistics, and converges faster than the Q-learning method when the number of channels is large.
\end{abstract}


\IEEEpeerreviewmaketitle

\section{Introduction}\label{sec:introduction}

According to CISCO~\cite{Cisco}, global mobile data traffic has experienced boosting growth, which will increase sevenfold between 2016 and 2021 with a compound annual growth rate (CAGR) of $46$ percent.
Spectrum extension is critical for future wireless communication networks to cope with this exponential data traffic growth.
However, radio spectrum is a costly and scarce resource and the use of radio spectrum is usually regulated by governmental entities.
The current shortage of radio spectrum makes it hard for wireless operators to obtain sufficient licensed bands with exclusive ownership.
On the other hand, experimental tests and investigations from both academia and industries show that the static spectrum allocation policy from Federal Communications Commission (FCC) causes the underutilization of allocated licensed bands~\cite{yin2012mining, Islam2008spectrum, mchenry2006chicago}.
To be specific, according to the report from Shared Spectrum Company (SSC), the utilization of most licensed spectrum bands is under $30\%$, and more than half of licensed spectrum bands are under $20\%$ utilized.
For example, even in the most crowded cities, like Washington DC and Chicago, only $38\%$ of licensed spectrum bands are occupied~\cite{SSC}.
These investigations and statistics reflect the fact that radio spectrum resources are under-utilized, which motivates FCC to reconsider the current static spectrum allocation policy, and employ dynamic spectrum access (DSA) to promote spectrum utilizations \cite{force2002spectrum}.

\IEEEpubidadjcol

A large quantity of public unlicensed frequency bands, such as the industrial, scientific, medical (ISM) band, and the unlicensed national information infrastructure (UNII) band, were freed up, on which any wireless devices are allowed to seek for spectrum extension using opportunistic spectrum access.
However, nowadays, these bands are becoming extremely crowded and over-utilized.
To address this issue, ultra-wideband millimeter wave (mmWave) bands have been exploited for unlicensed use, such as 24-24.25 GHz ISM bands and 28-30 GHz local multiple distribution service (LMDS) bands~\cite{pi2011introduction}.
Moreover, according to the recent spectrum allocation policy released by FCC, 14 GHz consecutive mmWave bands (57-71 GHz) can be used as license-exempt bands~\cite{FCC}.
Even though mmWave bands provide a lot of opportunities for radio spectrum access, it is important to note that the use of mmWave band usually requires sophisticated signal processing technologies and hardware.
To provide access opportunities on low frequency bands, FCC opened up licensed spectrum bands to encourage DSA of under-utilized spectrum~\cite{force2002spectrum}.
For example, in 2015, the FCC conducted an auction of 1695-1710 MHz, 1755-1780 MHz, and 2155-2180 MHz, also referred to as the Advanced Wireless Services-3 (AWS-3) band, on which DSA devices share the bands with federal systems, such as the federal meteorological-satellite systems~\cite{bhattarai2016overview}.
In addition to the AWS-3 band, the 3.5 GHz band has been opened up and utilized by the Spectrum Access System (SAS), in which different types of users will share the spectrum with different spectrum access priority, including Incumbent Access (IA), Priority Access (PA) and General Authorized Access (GAA)~\cite{federal2015report}.
Undoubtedly, opening up current licensed frequency bands could significantly improve the utilization of radio spectrum.
However, many challenges still exist.
For example, on opened licensed frequency bands, DSA users should always act as secondary users (SUs) with low priority. 
This means DSA users should provide protection to primary users (PUs) from harmful interference.
To achieve spectrum co-existence between DSA users and primary users, many strategies have been introduced so far, which can be classified into two main spectrum access mechanisms.
The first one is Listen-Before-Talk (LBT), also known as the interweaving scheme, in which a SU can access a frequency band only if it is detected to be available~\cite{liao2014listen}.
Although this scheme can effectively avoid strong interference to primary users, opportunities for DSA users to access shared frequency bands can be rather limited.
This is because under LBT the spectrum access depends completely on current spectrum sensing outcome.
In reality, due to the random nature of wireless environments, limited/no cooperation among SUs, and other practical factors spectrum sensing can never be perfect~\cite{fu2017throughput}.
This will cause false alarm or miss detection of PUs' activities leading SUs to make inappropriate decisions regarding channel access~\cite{fu2017throughput}.
The second spectrum access scheme is spectrum sharing, also known as the underlaying scheme~\cite{stotas2011enhancing}.
In this scheme, SUs coexist with the PUs on shared frequency bands, and adjust their transmit power level so that the accumulated interference experienced at PUs is less than a tolerable interference threshold.
This scheme requires a strong assumption that the channel state information between transmitters of SUs and the receivers of PUs are known as apirori in order to conduct power control.
However, in reality, it is usually difficult to obtain these channel state information without a central controller.
Even under the presence of a central controller, exchanging these channel state information may impose a heavy control overhead for the underlying network making it difficult to be implemented in practice.

In light of these challenges, machine learning-based approach has recently been introduced in the field of dynamic spectrum access due to its ability to adapt to dynamic unknown environments~\cite{Li2010Qlearning, Wang2018DRL, Naparstek2017DRL}. 
%
%
To be specific, with machine learning, spectrum access will be determined not only by current spectrum sensing outcome but also by the learning result from past spectrum status.
In this way, the negative impact of imperfect spectrum sensing could be greatly alleviated.
Additionally, machine learning can enable DSA devices to obtain accurate channel status and useful prediction/statistic information of channel status, such as the behavior of PUs and the load of other SUs, the spectrum access based on which could significantly reduce the collisions between SUs and PUs.

In this paper, we investigate machine learning approaches to obtain artificial intelligence-enabled (AI-enabled) spectrum access strategies for DSA networks.
To reduce the control overhead of the underlying DSA network, we incorporate the powerful machine learning technique, deep reinforcement learning (DRL), for SUs to learn ``appropriate'' spectrum access strategies in a distributed fashion assuming NO knowledge of the underlying system statistics.
Furthermore, a special type of recurrent neural network, called the reservoir computing (RC), is utilized to realize DRL by taking advantage of the underlying temporal correlation of the DSA network.
To be specific, DRL and RC based dynamic spectrum access scheme is developed to facilitate DSA systems to perform appropriate channel access, aiming at protecting primary users from harmful interference and avoid collisions with other SUs.
To the best of our knowledge, this is the first work to combine DRL and RC for DSA networks.
DRL aims to solve the large state space problem in traditional reinforcement learning~\cite{2013RNNdifficulty}.
Traditional reinforcement learning such as Q-learning converges slowly when the state space is large.
Meanwhile, DRL utilizes deep neural network to approximate the expected cumulative reward of the state-action pairs and accelerate the convergence time.
RC is a special type of recurrent neural network (RNN) that reduces the complexity of training significantly \cite{2009RCtoRNN}.
RNN is a powerful neural network that reasons the \emph{temporal correlation} of input sequences, but the difficulty of training RNN is a well-known trouble to many researchers \cite{2013RNNdifficulty, sak2014long}.
RC simplifies the training of RNN by only training the output weights and our previous work in this field including designing a low-complexity and high-efficiency RC-based symbol detector for MIMO-OFDM systems~\cite{MIMOOFDM2017}.
In our work, DQN and RC are combined to design dynamic spectrum access strategies for handling the large state space and taking advantage of the temporal correlation of the underlying wireless channel as well as the PU's activities.

It is important to note that DRL has been applied in \cite{Wang2018DRL, Naparstek2017DRL} for solving DSA problems.
%
%
In \cite{Wang2018DRL}, only one SU is assumed in the network with perfect spectrum sensing outcomes (that is, there is no error in spectrum sensing).
%
%
On the other hand, \cite{Naparstek2017DRL} assumes no PU in the network, therefore, the collision with PUs is not considered.
Furthermore, \cite{Wang2018DRL} combines DRL with multi-layer perceptron (MLP), which is incapable of learning the temporal correlation of spectrum sensing outcomes.
On the other hand, \cite{Naparstek2017DRL} combines DRL with RNN resulting in slow convergence rate, especially in large state space.

In this paper, to better reflect the reality, we assume that there are multiple PUs and multiple SUs in the DSA network by jointly considering effects of colliding with PUs or other SUs.
Furthermore, no centralized controller is assumed in our DSA network to allow fully distributed dynamic spectrum access strategies.
Imperfect spectrum sensing with unknown sensing error probabilities is also considered in the work.
Main contributions of this paper can be summarized in the following:
\begin{enumerate}
\item
A distributed dynamic spectrum access strategy for SUs of a DSA network is introduced based on the integration of DRL and RC. The strategy takes SUs' imperfect spectrum sensing outcomes into account and enables SUs to conduct spectrum access in a fully distributed fashion. 
\item
Extensive performance evaluation of the introduced dynamic spectrum access strategy has been conducted to show that our
machine learning based spectrum access strategy could quickly learn PUs' activities and significantly reduce the chance of collisions with PUs and SUs in a realistic environment.
\item
Compared with the myopic scheme which assumes system statistic, our scheme can outperform the myopic strategy in both the single-SU case and the multiple-SU case.
Compared with Q-learning, our scheme demonstrate quick convergence rate with better performance.
Compared with DRL+MLP, our DRL+RC strategy can take advantage of the underlying temporal correlation of the sensing result to yield substantial performance improvement. 
\end{enumerate}

\section{deep reinforcement learning and reservoir computing}
When system dynamics are unknown and observations of the system are not entirely accurate, learning-based methods can adapt to the unknown environment from the partial observation of the system by improving performance on a specific task.
Usually, a SU does not know the behaviors of PUs and its spectrum sensor may make a mistake in determining the spectrum holes.
Therefore, machine learning is suitable for designing dynamic spectrum access strategies when the system statistics is unknown.
In this work, we adopt DRL and RC, in particular, deep Q Network (DQN) \cite{mnih2015DRL} and echo state network (ESN) \cite{jaeger2001echo}.

DRL has attracted much attention in recent years because it enables reinforcement learning to efficiently learn in a very large state and action spaces by providing a good approximation of Q-value.
Due to the difficulties of training recurrent neural network (RNN), reservoir computing is a new paradigm of RNN training that is simple but powerful.
We use a special type of reservoir computing, echo state network (ESN), as the Q-network of DQN.

\subsection{Reinforcement Learning}
Reinforcement learning is an important type of machine learning where an agent learns how to behave in an environment to maximize the cumulative reward.
It lies between supervised and unsupervised learning.
The agent is not told which action to take, which is different from the ground truth labels in supervised learning.
Instead, the agent must explore the environment to accumulate knowledge of the environment and exploit its accumulated knowledge to take the best action it can.
The agent gradually develops a policy that maps from the observed state to the action.
In the distributive dynamic spectrum access work of our interests, each SU is an agent, sensing decision is the action, sensing outcome is the observed state, and sensing strategy is the developed policy.

The most wide-used training technique for reinforcement learning is Q-learning \cite{1992QLearning}.
This technique is a model-free approach that learns the policy directly through interactions with the environment without estimating a model of the environment.
Q-learning is a value iteration approach to find the Q-value $Q^{\pi}(s,a)$ of each state $s$ and action $a$ pairs, which represents the expected value of the cumulative reward when taking action $a$ in the initial state $s$ and then following policy $\pi$ thereafter.
$Q^{\pi^{*}}(s,a)$ is the Q-value with the optimal policy, and the optimal policy can be derived as
\begin{equation}
\pi^{*}(s) = \argmax_{a} Q^{\pi^{*}}(s,a)
\label{eq:Optimal policy of QLearning}
\end{equation}
Assume that the initial state is $s_t$, an agent takes the action $a_t$ following current policy $\pi$ and gains a reward $r_{t+1}$, then enters to the next state $s_{t+1}$.
The objective is to maximize the cumulative discounted reward $R$.
\begin{equation}
R = \sum_{t=1}^{\infty} \gamma^{t-1} r_{t+1}^l
\end{equation}
where $\gamma \in [0,1]$ is the discounted rate of the reward.
The online update rule of Q-values is given below
\begin{equation}
\begin{split}
Q(s_t,a_t) &\leftarrow Q(s_t,a_t) \\
&+ \alpha[ r_{t+1}
+ \gamma \max_{a_{t+1}}{Q(s_{t+1},a_{t+1})} - Q(s_t,a_t)]
\label{eq:QLearning Update}
\end{split}
\end{equation}
where $\alpha \in (0,1) $ is the learning rate and $\gamma \in [0,1]$ is the discounted rate of the reward.
Once the Q-value is updated, the policy $\pi$ is also updated using $\epsilon$-greedy method \cite{egreedy}.
\begin{equation}
a_{t+1} = \begin{cases}
\argmax_{a} Q(s,a), & \text{with probability $1-\epsilon$} \\
\text{Choose random action,} & \text{with probability $\epsilon$}
\end{cases}
\label{eq:epsilon greedy}
\end{equation}
where $\epsilon$ is a small number between 0 and 1.
The Q-value and the policy updates iteratively to gradually converge to the optimal policy.

\subsection{Deep Reinforcement Learning}
Q-learning performs well for small-scale models but performs poorly for large-scale models.
The reason is that the training algorithm of Q-learning iteratively updates the Q-table.
As the number of possible states increases, the large Q-table size makes training difficult or even impossible \cite{2013RNNdifficulty}.

Due to the difficulties of updating every element in Q-table for a large-scale model, DRL exploits the powerful deep neural network to approximate the Q-value.
In 2013, DeepMind team developed a Deep Q-network (DQN) to challenge seven Atari 2600 games and won a human expert on three of the games \cite{2013atari}.
In our work, the size of the state space grows exponentially with the number of the channels.
Each channel is occupied by a PU and each PU has two possible states: \textit{Active} state or \textit{Inactive} state, so the state size is $2^N$ for $N$ channels.
This motivates us to use DQN to learn the dynamic spectrum access strategy in an unknown dynamic system.

\begin{figure}[t]
\centering
	\includegraphics[width=.9\linewidth]{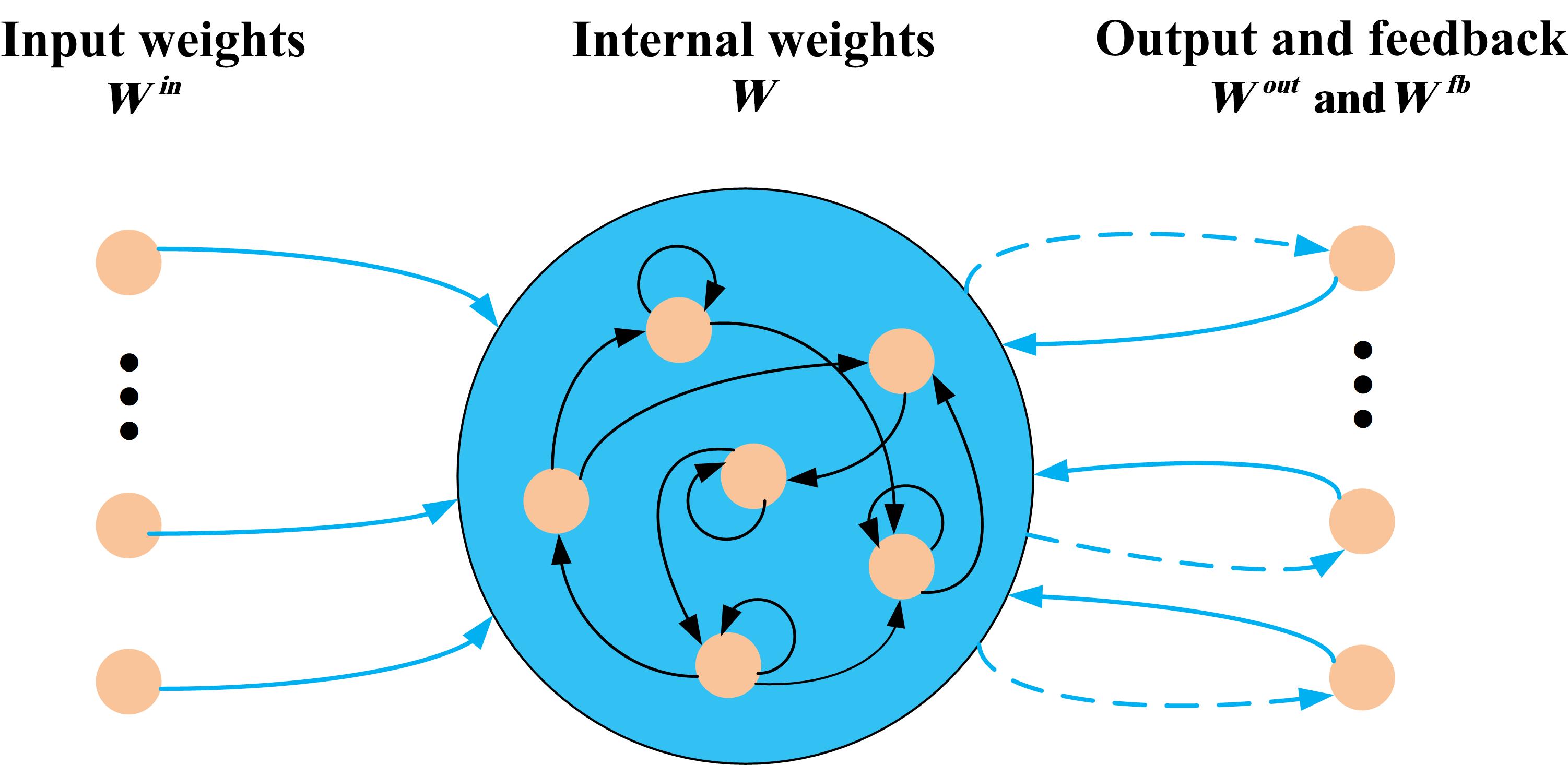}
	\caption{Reservoir Computing.}
	\label{fig:reservoir_computing}
\end{figure}

\subsection{Reservoir Computing}
Recurrent neural network (RNN) is a powerful network that is capable of learning the temporal behavior for a time sequence.
Compared with the MLP, RNN has feedback connections that allow signals to travel in both directions, which exhibits a dynamic behavior.
Also, RNN has memory that stores the previous neuron activations.
RNN are widely used in the natural language processing because understanding a sentence highly requires temporal relationship between words~\cite{mikolov2010recurrent, sak2014long}.
Although RNN is more powerful than the feedforward neural network (FFNN) in terms of the expressiveness, RNN is more challenging to train~\cite{2013RNNdifficulty, bengio1994learning}.
The training algorithms for RNN have the vanishing and the exploding gradient problem because the gradient values calculated in long temporal range may become too large or too small, which makes the time for training RNN very long.
The aforementioned large state size of channels in DSA further makes training RNN more difficult since we have to model a large-scale dynamic system.
Therefore, we need to find the alternative way of training RNN that have faster convergence.

Due to the difficulties of training RNN, RC is introduced to simplify the training with comparable performance~\cite{2009RCtoRNN}.
The reservoir is a randomly generated RNN, and it has linear connections to input units and output units, which is shown in Fig.~\ref{fig:reservoir_computing}.
During the training, only the output weights are trained, the weights inside the reservoir and the inputs weights are fixed \cite{jaeger2002tutorial}.
The fixed recurrent connections of the reservoir provide a high dimensional dynamics that is able to  create all possible spatial and temporal combinations of the input history, which are analogous to cortical dynamics in human brain \cite{enel2016reservoir}.
The learned output weights determine the best linear combination of the reservoir's state and the input signal to perform the desired task.
This approach largely reduces the computation time because only the read out weights are trained.
ESN is the most popular type of reservoir computing system \cite{jaeger2007echo, kian2018DFR, MIMOOFDM2017, Mingzhe2017ESN}.
In this work, we adopt ESN as the Q-network in the DQN.

\begin{figure}[t]
\centering
	\includegraphics[width=1.0\linewidth]{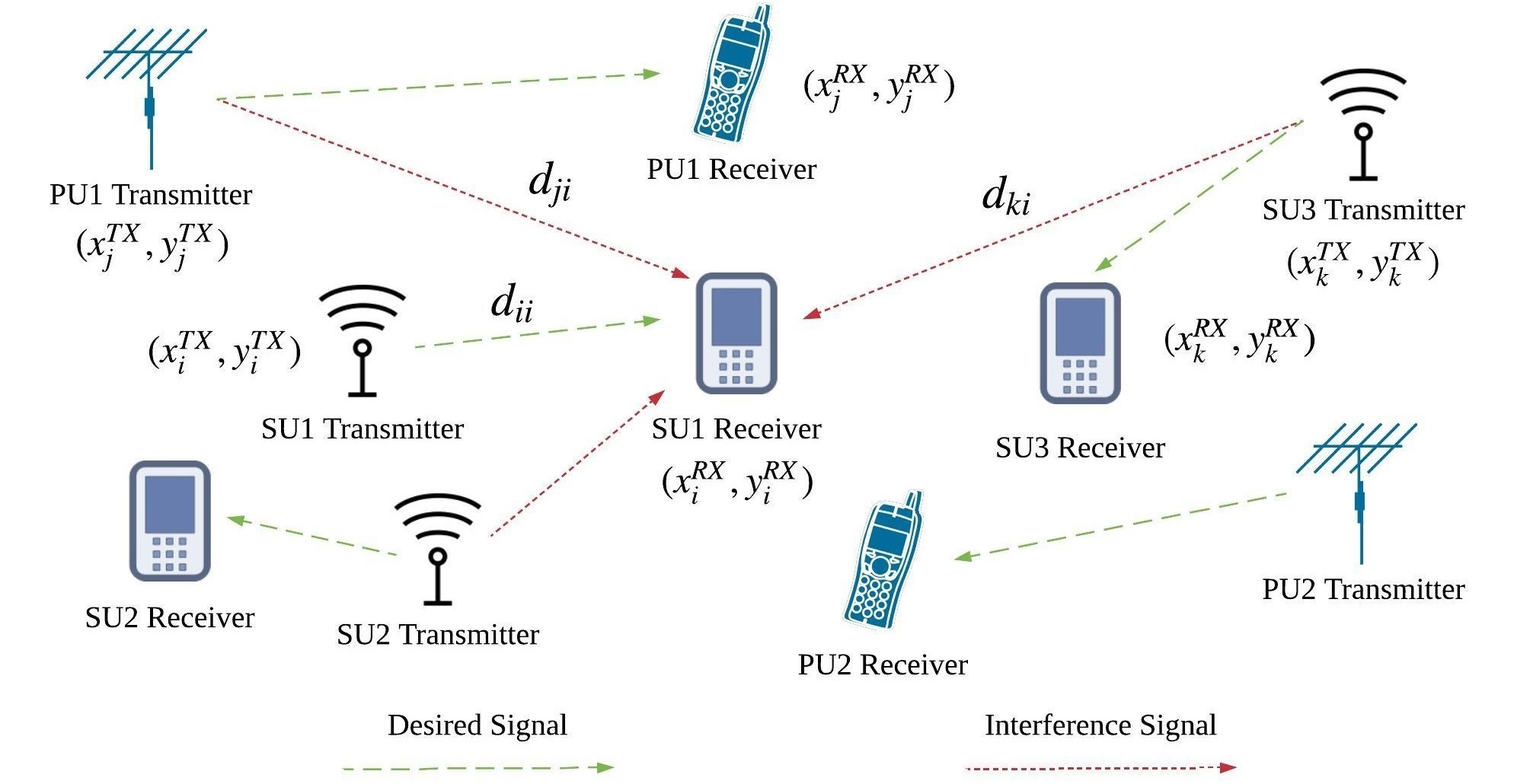}
	\caption{System Model.}
	\label{fig:framework}
\end{figure}

\section{System Model}
\label{System model}
We consider a DSA scenario in which primary network consisting of $N$ PUs coexist with DSA secondary network with $L$ SUs.
It is assume that there are totally $N$ wireless channels so that each PU transmit on one unique wireless channel to avoid interference among PUs.
Furthermore, we assume that each PU will broadcast warning signals to SUs if the corresponding PU's signal has been collided.
To be specific, the warning signal contains information related to which channel has been collided so that the corresponding transmitting SUs on that channel are aware of the collision.
This is the only control information from the PU to the SU to enable the protection for the PU.
Our proposed method outperforms myopic method that is required to know the system statistics, and converges faster than Q-learning method when the number of channels is large.
All wireless channels are shared by all SUs, and a SU will select proper wireless channels to access according to its spectrum access strategy.
Moreover, on a wireless channel, a SU should protect the PU, which is authorized to access the channel, from harmful interference.

As shown in Fig.~\ref{fig:framework}, on the X-Y two-dimensional coordinates, $\left( {x_i^{TX},y_i^{TX}} \right)$, $\left( {x_i^{RX},y_i^{RX}} \right)$, $\left( {x_j^{TX},y_j^{TX}} \right)$, and $\left( {x_j^{RX},y_j^{RX}} \right)$ respectively represent the position coordinates of the transmitter of SU $i$, the receiver of SU $i$, the transmitter of PU $j$, and the receiver of PU $j$, where $i \in \{ 1,2, ...,L \}$ and $j \in \{ 1,2, ...,N \}$.
Accordingly, the distance of a desired signal link can be calculated by ${d_{ii}} = \sqrt {{{\left( {x_i^{TX} - x_i^{RX}} \right)}^2} + {{\left( {y_i^{TX} - y_i^{RX}} \right)}^2}}$.
Moreover, the propagation distance of interference signal from and other SUs are given by ${d_{ji}} = \sqrt {{{\left( {x_j^{TX} - x_i^{RX}} \right)}^2} + {{\left( {y_j^{TX} - y_i^{RX}} \right)}^2}}$ and ${d_{ki}} = \sqrt {{{\left( {x_k^{TX} - x_i^{RX}} \right)}^2} + {{\left( {y_k^{TX} - y_i^{RX}} \right)}^2}}$, respectively, where $k \in \{ 1,2, ...,L \}$ and $k \ne i$.
The interference only happens when PU $j$ and SU $k$ are using the same wireless channel that SU $i$ is using.

With the propagation distance of $d$, we adopt the WINNER II channel model \cite{meinila2009winner} to calculate the path loss of the desired signal using

\begin{equation}
PL\left( {d,{f_c}} \right) = \overline {PL}  + A_W \cdot {\log _{10}}\left( {d [m]} \right) + B_W \cdot {\log _{10}}\left( {\frac{{{f_c}[GHz]}}{5}} \right),
\label{equ:winnerII}
\end{equation}
where $f_c$ is the carrier frequency of wireless channels.
$\overline {PL}$, $A_W$, and $B_W$ denote the path loss at a reference distance, path loss exponent, and path loss frequency dependence, respectively.
Accordingly, the path loss of desired signal $PL\left( {d_{ii},{f_c}} \right)$ and interference signals $PL\left( {d_{ji},{f_c}} \right)$ and $PL\left( {d_{ki},{f_c}} \right)$ can be obtained.
It is assumed that a strong Line-of-Sight (LOS) path exists between a transmitter and a receiver, therefore the Rician channel model is employed to derive channel model, which can be expressed as
\begin{equation}
h= \sqrt {\frac{\kappa }{{\kappa  + 1}}} \sigma {e^{j\theta }} + \sqrt {\frac{1}{{\kappa  + 1}}} CN\left( {0,{\sigma ^2}} \right),
\label{equ:rician}
\end{equation}
where ${\sigma ^2} = {10^{ - \frac{{\overline {PL}  + A \cdot {\log _{10}}\left( {d[m]} \right) +  B \cdot {\log _{10}}\left( {\frac{{{f_c}[GHz]}}{5}} \right)}}{{10}}}}$ is determined by path loss, and $\kappa$ is $K$-factor which indicate the ratio between receiver signal power of a LOS path and scattered paths.
$\theta \sim U(0, 1)$ is the phase of arrival signals on LOS path, which takes value from the uniform distribution between 0 and 1, and $CN\left( \cdot \right)$ represents a circularly symmetric complex Gaussian random variable.

Thus, the signal to interference plus noise ratio (SINR) of received signals at the receiver of SU $i$ can be derived by

\begin{equation}
SINR_{i} = \frac{{{p_{ij}} \cdot {{\left| {{h_{ii}}} \right|}^2}}}{{{p_{jj}} \cdot {{\left| {{h_{ji}}} \right|}^2} + \sum\limits_{k = 1, k \neq i}^L {{p_{kj}} \cdot {{\left| {{h_{ki}}} \right|}^2}} {\text{ + }}B \cdot {N_0}}},
\label{equ:SINR}
\end{equation}
where $p_{jj}$, $p_{ij}$ and $p_{kj}$ represent the transmit power of PU $j$, SU $i$, and SU $k$ on the $j$th wireless channel, respectively.
${\left| {{h_{ii}}} \right|}^2$, ${\left| {{h_{ji}}} \right|}^2$, and ${\left| {{h_{ki}}} \right|}^2$ respectively denote the channel gain of the links between the transmitter $i$ and the receiver $i$, between the transmitter $j$ and receiver $i$, and between the transmitter $k$ and receiver $i$. $B$ and $N_0$ stand for channel bandwidth and noise spectral density, respectively.
With the received SINR of SU $i$, the channel capacity could be obtained by ${C_i} = B \cdot {\log _2}\left( {1 + SIN{R_i}/\Gamma} \right)$.
Here, $\Gamma$ is the SINR gap that defines the gap between the channel capacity and a practical modulation and coding scheme (MCS).

\section{DQN-based Spectrum Access Strategy}
\begin{figure}[t]
\centering
	\includegraphics[width=.7\linewidth]{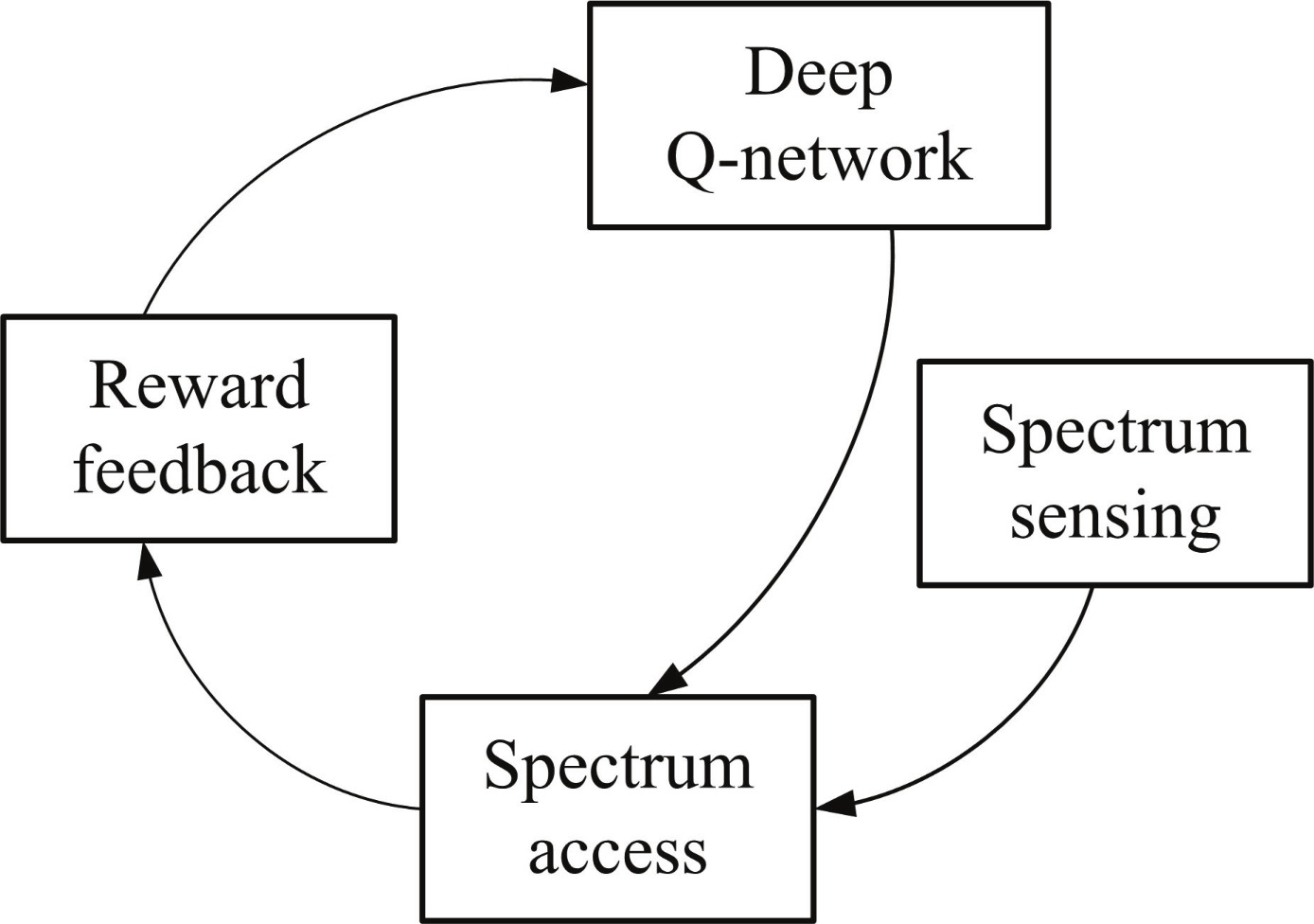}
	\caption{Learning procedure in DSA.}
	\label{fig:learning_procedure}
\end{figure}

In this section, we formulate the distributive dynamic spectrum access as a reinforcement learning problem.
We define the agent, state, action, reward, and policy in dynamic spectrum access environment.
The learning procedure is shown in Fig.~\ref{fig:learning_procedure}
.
It can be seen that spectrum access strategies are determined by the results of deep Q-network and current spectrum sensing.
According to the spectrum access strategies, SUs access wireless channels to carry out data transmissions.
Then, SU receivers feedback reward based on actual wireless transmission quality, which will be stored by SU transmitters and used as training data of DQN+RC to update spectrum access strategies.
The aforementioned learning procedure will be carried out periodically to tackle the variations of wireless environments.

\begin{figure}[t]
\centering
	\includegraphics[width=.9\linewidth]{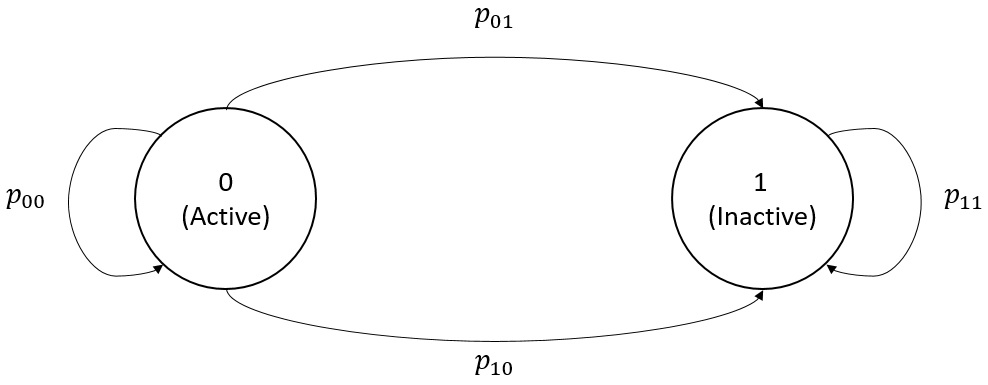}
	\caption{Two-state Markov chain.}
	\label{fig:two_state_markov_chain}
\end{figure}

There are a set of $\{1,2,...,N\}$ orthogonal channels and a set of $\{1,2,...,L\}$ SUs.
Each channel is occupied by a PU and each PU may be in one of the two states: \textit{Inactive} (1) or \textit{Active} (0).
The PU with \textit{Inactive} state means SU is allowed to access the corresponding channel, and the PU with \textit{Active} state means SU cannot access the corresponding channel because the PU is accessing it.
The dynamic of each PU's activity is described as a two-state Markov chain as shown in Fig.~\ref{fig:two_state_markov_chain}.
The transition probability of the two-state Markov chain on the $n$th channel is denoted as
\begin{equation}
P_n = \left[
 \begin{matrix}
   p_{00}^n & p_{01}^n \\
   p_{10}^n & p_{11}^n
  \end{matrix}
  \right]
\label{equ:Markov}
\end{equation}
where $p_{ij} = \text{Pr}\{ \text{next state is j } | \text{current state is i} \}$ ($i,j \in \{0,1\}$).

At the beginning of each time slot, every SU performs spectrum sensing on all $N$ channels to detect the channel states.
Let the sensing results at time slot $t$ be
\begin{equation}
S(t) = [S^1(t),..., S^L(t)]
\end{equation}
where $S^l(t)$ is a $N$-dimensional vector $[s_1^l(t),...,s_N^l(t)]^T$, and $s_n^l(t) \in \{ 0, 1 \}$ is the sensed state of the $l$th SU on the $n$th channel.
The spectrum sensing detector is not perfect, so $s_n^l(t)$ may contain error.
Let the sensing error probability of the $l$th SU on the $n$th channel be $E_n^l$.
\begin{equation}
\text{Pr} \{ T_n(t)  = s_n^l(t) \} = 1 - E_n^l
\end{equation}
where $T_n(t)$ is the true state of the $n$th channel.
The transition probabilities and the sensing error probabilities of the channels are both unknown to SUs.
The only known information for the $l$th SU is $S^l(t)$ representing the observed state in the environment and the input of the DQN.

After performing spectrum sensing, each SU decides to access at most one channel or stay idle based on the sensing result.
The action of the $l$th SU is denoted as
\begin{equation}
a^{l}(t) \in \{0,...,N\}
\end{equation}
where $a^{l}(t)=n$($n>0$) means the $l$th SU decides to access the $n$th channel at time slot $t$, and $a^{l}(t)=0$ means the $l$th SU decides not to access any channel at time slot $t$.

After a SU transmitter accesses a channel, channel states are changed according to its Markov chains.
Then the corresponding SU receiver feedbacks the received SINR to the transmitter.
As discussed in Section~\ref{System model}, the interference will be based on the locations of PUs and SUs.
We consider four cases for setting up the reward function:
\begin{enumerate}
\item
A SU accesses a channel that no PU or other SUs are currently using:
In this case, the SU does not experience any interference.
The achievable data transmission rate, $\log_2(1+SINR/\Gamma)$, is used as the reward function.
\item
A SU accesses a channel that is currently occupied by a PU:
In this case, a collision with PU occurs. 
According to the system model described in Section~\ref{System model} the PU will broadcast warning signals to all SUs.
Therefore, the corresponding transmitting SUs will be aware of this collision and we set a negative reward of $-C$ ($C>0$) as an outcome of receiving the warning signal.
%
%
\item
More than two SUs access the available channel: In this case, collisions happen among SUs.
Similarly, the achievable data transmission rate, $\log_2(1+SINR/\Gamma)$, is used as the reward function.
\item
A SU decides not to access any channel, and we set the reward to be zero since $\log_2(1+SINR/\Gamma)=0$.
\end{enumerate}

As an outcome, the reward function of the $l$th SU on the $n$th channel can be expressed as
\begin{equation}
r^l(t+1) = \begin{cases}
-C, & \text{Interference with PU} \\
\log_2(1+SINR_n^l/\Gamma), & \text{otherwise}
\end{cases}
\label{eqn:reward}
\end{equation}

The dynamic spectrum access strategy is distributive so that the information of sensing results and accessing decisions are not shared between SUs.
Each SU has its DQN to make decisions of channel access independently, and the only input to each SU's DQN is sensing results implemented by its sensor.
SUs do not know the transition probabilities of channel states and the probabilities of sensing errors. They can only learn how to access the channel through the received SINR that they get after accessing, based on which their access strategies are developed to maximize their own cumulative discounted reward $R^l$:
\begin{equation}
R^l = \sum_{t=1}^{\infty} \gamma^{t-1} r^l(t+1)
\end{equation}
where $\gamma \in [0,1]$ denotes the discounted rate. $r^l(t+1)$ is defined in~\eqref{eqn:reward}.
The training method for all DQNs is shown as Algorithm~\ref{alg:DQN_training}.

\begin{figure*}[t]
\centering
	\includegraphics[width=.73\linewidth]{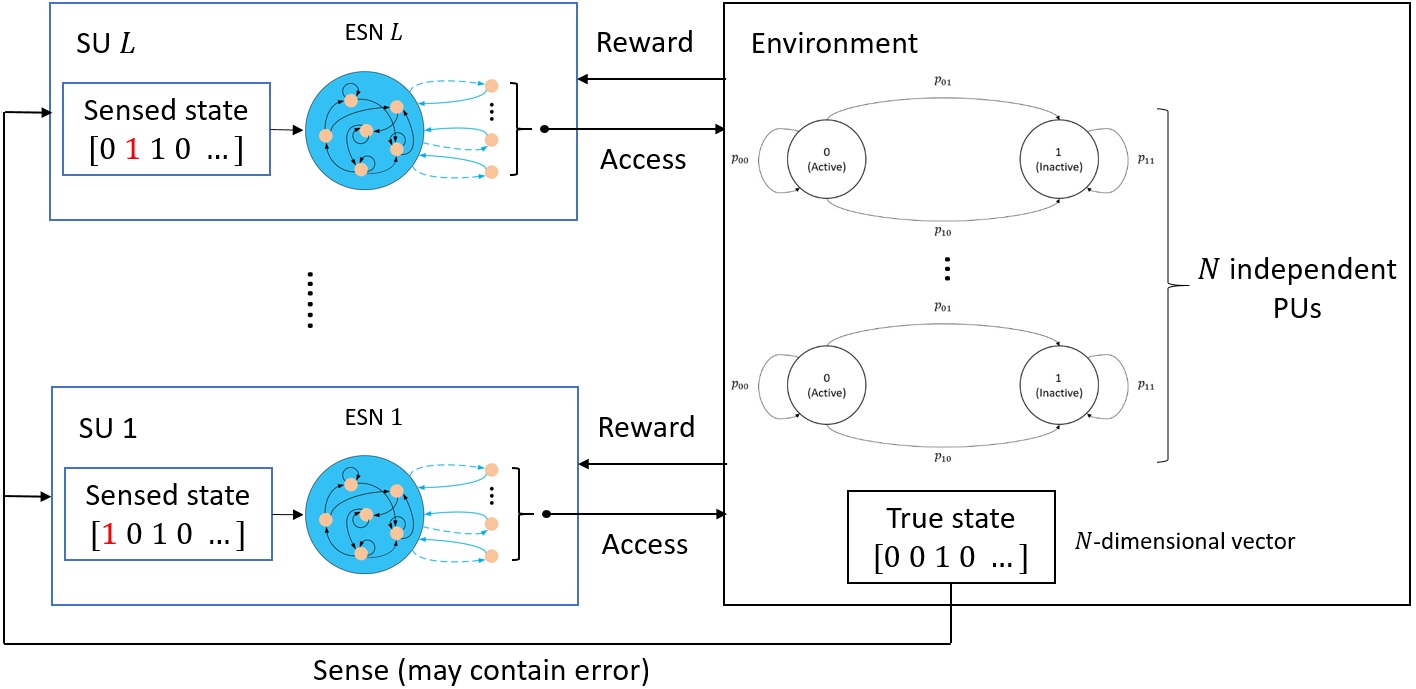}
	\caption{Architecture for distributive deep reinforcement learning.}
	\label{fig:architecture}
\end{figure*}

\begin{algorithm}
\caption{DQN training process}
\label{alg:DQN_training}
\begin{algorithmic}
\STATE {
1) Initialize $\text{DQN}_{\text{t}}^l$ and $\text{DQN}_{\text{e}}^l$ with the same structure and initial weights for each SU $l$.}\\

2) Each SU $l$ inputs $S^l(t)$ to $\text{DQN}_{\text{e}}^l$ and chooses the action $a^l(t)$ based on Equation~\ref{eq:epsilon greedy}.

3) Each SU $l$ gets a reward $r_{t+1}^l$, and observes the next state $S^l(t+1)$.

4) Store $S^l(t), a^l(t), r^l(t+1), S^l(t+1)$ and replace $S^l(t)$ with $S^l(t+1)$.

5) Repeats step 2 to step 4 for $T$ times at $T$ slots.

6) Given the stored $T$ sequences, $S^l(t), a^l(t), r^l(t+1), S^l(t+1)$, input $S^l(t)$ and $S^l(t+1)$ to $\text{DQN}_{\text{e}}^l$ and $\text{DQN}_{\text{t}}^l$ to generate $\text{Q}_{\text{e}}^l(S^l(t), a)$ and $\text{Q}_{\text{t}}^l(S^l(t+1), a)$, respectively. Then update the $\text{DQN}_{\text{e}}^l(a)$ to minimize the mean-square-error of the following function:
\begin{equation*}
[r^l(t+1) + \gamma \max_{a}{\text{Q}_{\text{t}}^l(S^l(t+1),a)} - \text{Q}_{\text{e}}(S^l(t),a^l(t))]^2
\end{equation*}

8) Replace $\text{DQN}_{\text{t}}^l$ with $\text{DQN}_{\text{e}}^l$:

\end{algorithmic}
\end{algorithm}
The architecture of the proposed distributive dynamic spectrum access using deep reinforcement learning and reservoir computing is shown in Fig.~\ref{fig:architecture}.

\section{Performance Evaluation}
In this section, we present the experiment results for the distributive DQN-based dynamic spectrum access.
We first randomly choose the locations of SUs and PUs in a 150m$\times$150m square.
The distance between a SU transmitter and the corresponding receiver is randomly chosen from 20m and 40m.
As discussed in Section 3, the WINNER II model in~\eqref{equ:winnerII} and the Rician model in~\eqref{equ:rician} are employed to calculate the path loss and our derived channel model, respectively.
For the WINNER II model, we set  $f_c=5$GHz, $\overline {PL}=41$, $A_W=22.7$, and $B_W=20$.
For the Rician model, we set $\kappa$ to 8, and $\sigma^2$ is determined by the path loss obtained from the WINNER II model.
Then the received SINR at a SU's receiver is given by~\eqref{equ:SINR}, where the bandwidth $B$ is 1MHz, the noise spectral density $N_0$ is $10^{-14.7}$(mW/Hz), the transmit power of one SU and is 20mW, and the transmit power of one PU is 40mW.
All system parameters used in the channel model are listed in Table~\ref{tab:channel_model_parameters}.

As described in Section 4, the dynamics of PUs are modelled as independent two-state Markov chains with states: \textit{Inactive} (1) or \textit{Active} (0).
To initialize each Markov chain as shown in~\eqref{equ:Markov}, we randomly choose $p_{11}$ and $p_{00}$ from a uniform distribution over $[0.7, 1]$ and $[0, 0.3]$ respectively for every channel.
Then $p_{10} = 1 - p_{11}$ and $p_{01} = 1 - p_{00}$ can be calculated accordingly.
The reason for choosing these ranges for $p_{11}$ and $p_{00}$ is that the utilization of most licensed spectrum bands is low, so the possible value of $p_{00}$ should be low, and the possible value of $p_{11}$ should be high.

\begin{table}
	\centering
	\caption{Channel model paramters}
	\label{tab:channel_model_parameters}  
	\begin{tabular}{|c|c|}
	\hline
	SU transmit power & 20mW \\ \hline
    PU transmit power & 40mW \\ \hline
    $f_c$ & 5GHz \\ \hline
    $\overline {PL}$ & 41 \\ \hline
    $A_W$ & 22.7 \\ \hline
    $B_W$ & 20 \\ \hline
    $\kappa$ & 8 \\ \hline
    $B$ & 1MHz \\ \hline
    $N_0$ & $10^{-14.7}$(mW/Hz) \\ \hline
    \end{tabular}
\end{table}

To demonstrate the effectiveness of our proposed learning method, we employ the myopic and the Q-learning methods as referred methods, which will be compared with learning-based method.
In the myopic method, the channel with the maximum immediate reward will be selected~\cite{Liu2010Myopic,Li2012Myopic}.
To calculate the expected immediate reward, the myopic method needs the information of the transition probabilities of channels and the probabilities of sensing error.
Therefore, we regard the myopic method as a baseline which has more known information than other learning methods, such as DQN and Q-learing.
In the myopic method, $l$th SU firstly calculates channel state probabilities with $G_n(t) = \text{Pr}\{\text{the $n$th channel state at time slot $t$}=1 \}$.
\begin{equation}
G_n(t) = s_n^l(t) \cdot (1-E_n^l) + (1-s_n^l(t)) \cdot E_n^l
\end{equation}
where $s_n^l(t) \in [0, 1]$ is the sensing result of $l$th SU on $n$th channel. $E_n^l$ represents the sensing error probability of $l$th SU on $n$th channel.
Then the myopic method calculates the expected immediate reward $R_n(t)$ of $n$th channel.
\begin{equation}
\begin{aligned}
&R_n(t) = G_n(t) \cdot [ p_{10}^n \cdot (-C) + p_{11}^n \cdot \log_2(1+SINR_n^l/\Gamma)] \\
&+ (1 - G_n(t)) \cdot [ p_{00}^n \cdot (-C) + p_{01}^n \cdot \log_2(1+SINR_n^l/\Gamma)]
\end{aligned}
\end{equation}
Thus, with the myopic method, a SU will access the channel with maximal $R_n(t)$.

On the other hand, Q-learning algorithms have widely been adopted to solve multi-user dynamic spectrum access problems in the existing literatures~\cite{2010RL,Morozs2016Qlearning}.
The main drawback of Q-learning is that it is not able to handle a large state size.
Since DQN approximates the mapping function from the state space to the Q-value instead of frequently updating Q-table, it has faster convergence than Q-learning when the state space is large.

\subsection{Single SU}
\begin{figure}[ht]
\centering
	\includegraphics[width=.75\linewidth]{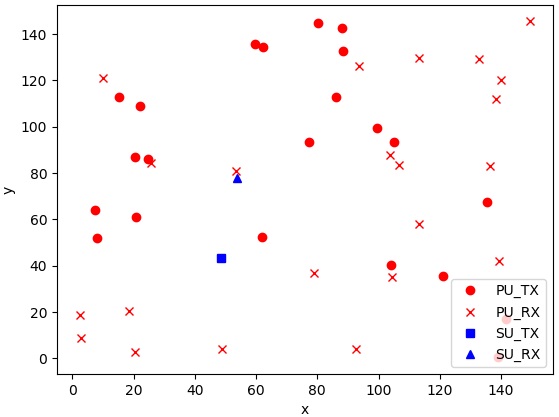}
	\caption{DSA Network Geometry (No. of channel = 22 and No. of SU = 1)}
	\label{fig:geometry_channel22_su1}
\end{figure}

\begin{figure}[ht]
\centering
	\includegraphics[width=.95\linewidth]{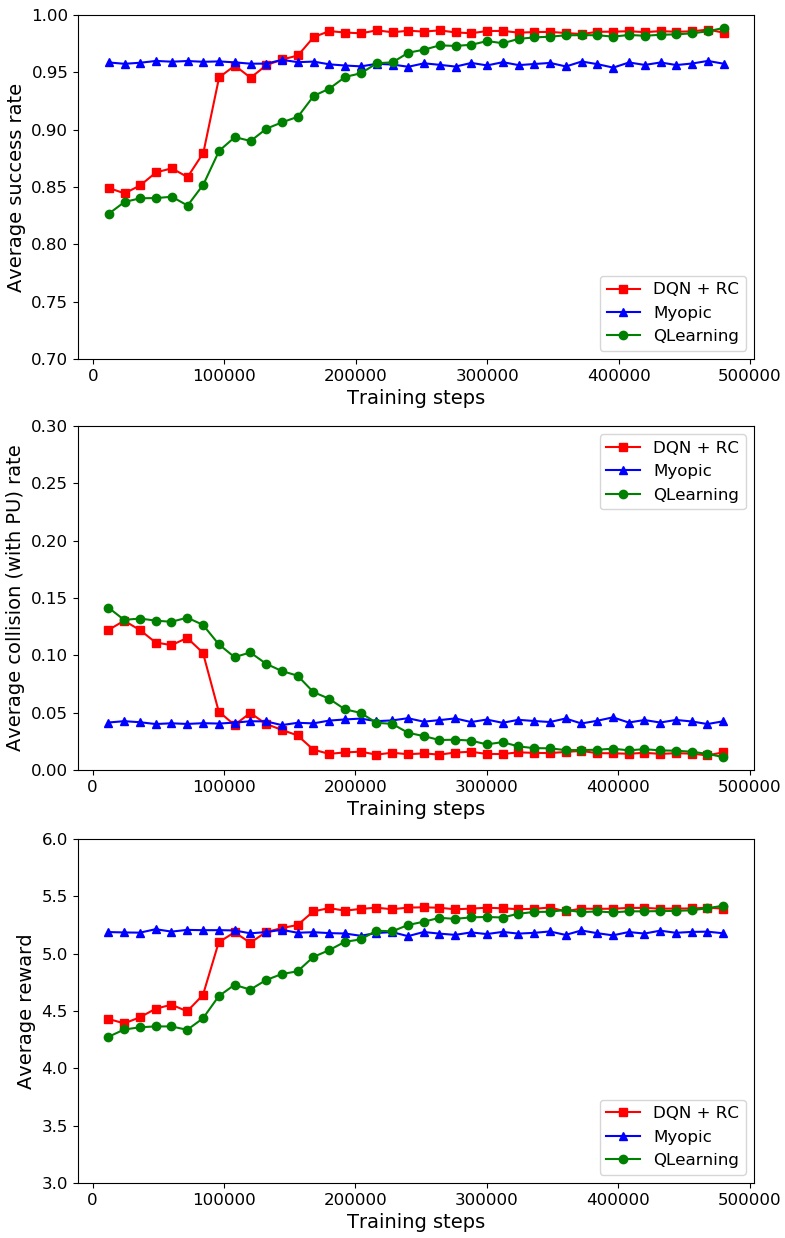}
	\caption{The average success number, collision number with PUs, and reward with the training time (No. of channel = 22 and No. of SU = 1)}
	\label{fig:result_channel_22_su_1}
\end{figure}

In this section, we design experiments to measure the convergence speed of our proposed learning method.
In the experiment, we treat the myopic and Q-learning method as the reference.
We set the number of SU and channels to 1 and 22, respectively.
Note that the number of PUs is the same as that of channels.
Furthermore, the negative reward for colliding with a PU is set to be $-2$ and the number of neurons of the RC is set to be $64$.
To ensure a fair comparison, we train both the DQN+RC and Q-learning with the same learning rate of $0.01$.
The locations of PUs and SUs in the underlying DSA network are shown in Fig.~\ref{fig:geometry_channel22_su1}.
In Fig.~\ref{fig:result_channel_22_su_1}, we show the average success rate (no collision with PU or SU), the average collision rate with PUs, and the average reward versus the training time.
In each iteration, we collect $4000$ training sequences ($T=4000$ in step 5 of Algorithm~\ref{alg:DQN_training}) and calculate the arithmetic mean of the success rate, the collision rate, and the reward.
After that we use the trained DQN to determine the access policy for the following iteration.
Although Q-learning and DQN+RC have similar performance when they converge, it can be seen that Q-learning converges much slower than DQN+RC when the number of channels is large.
This is because Q-learning needs intensive Q-table updates, and DQN+RC approximates the Q-value using RC, so DQN+RC converges faster if the approximation is accurate.
In addition, the results show that DQN+RC and Q-learning can outperform the myopic method even though the myopic method treats transition probabilities of channels and probabilities of sensing error as known information.
Therefore, the evaluation result directly suggests that learning-based methods can better adapt to the network dynamics than the myopic method, which assumes more apriori information.
To sum up, our proposed DQN+RC has the advantages of faster convergence speed when the number of channels is large and better performance than the myopic method.

\subsection{Multiple SUs}
\begin{figure}[ht]
\centering
	\includegraphics[width=.75\linewidth]{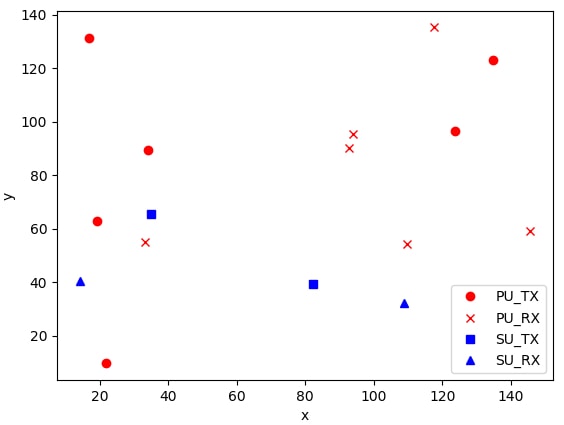}
	\caption{DSA Network Geometry (No. of channel = 6 and No. of SU = 2)}
	\label{fig:geometry_channel6_su2_punish_-2}
\end{figure}

\begin{figure}[ht]
\centering
	\includegraphics[width=.95\linewidth]{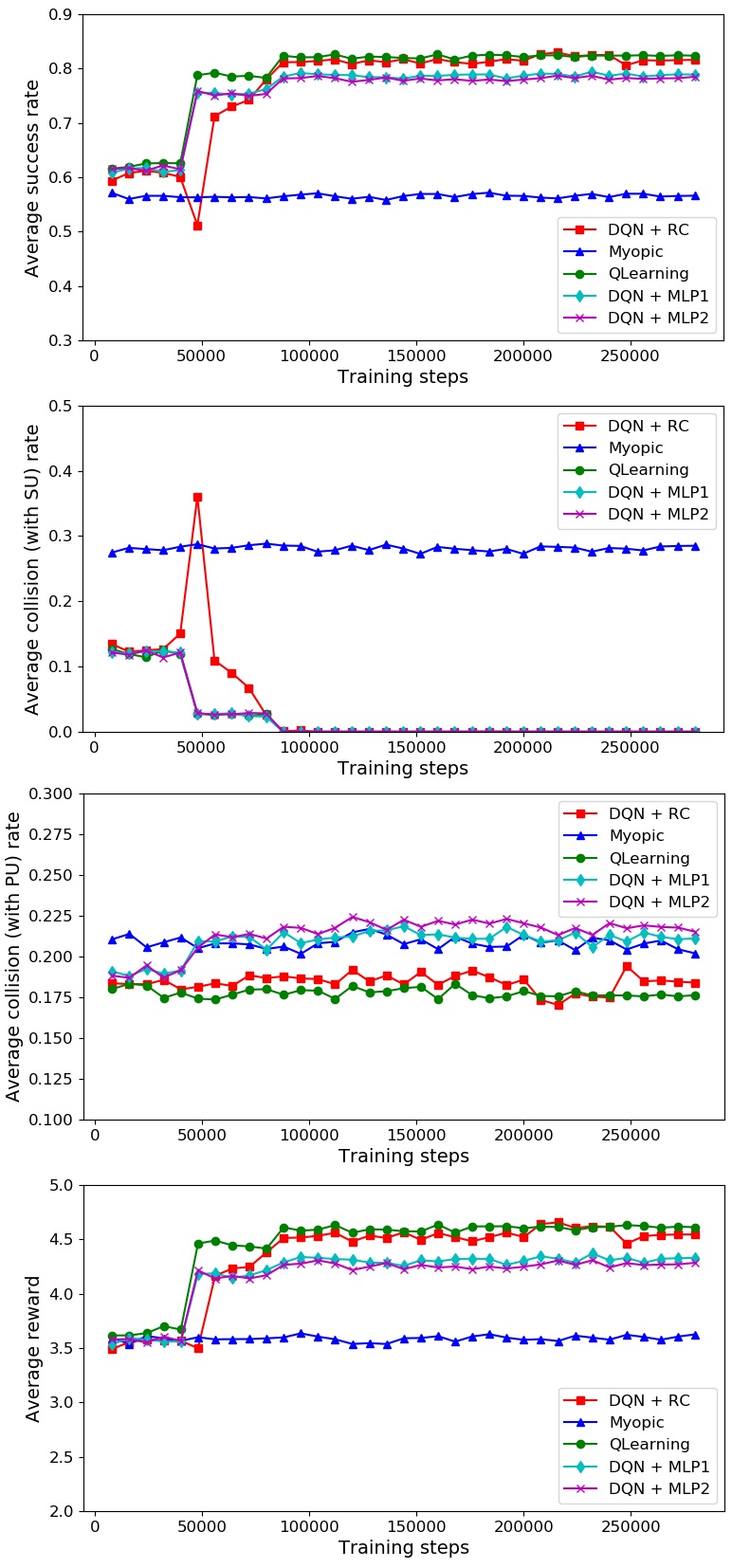}
	\caption{The average success number, collision number with PUs, collision number with SUs, and reward with the training time (No. of channel = 6 and No. of SU = 2)}
	\label{fig:result_channel6_su2_punish_-2}
\end{figure}

Fig.~\ref{fig:result_channel6_su2_punish_-2} shows the average success rate, the average collision rate with PUs and SUs, and the average reward versus the training time for a DSA network having multiple SUs. 
The network geometry of the underlying DSA network is shown most clearly in Fig.~\ref{fig:geometry_channel6_su2_punish_-2}.
Since SUs may collide with each other, each SU also needs to learn the access strategies of other SUs.
We set the number of channels and SUs to be $6$ and $2$, respectively.
The negative reward for colliding with PU is set to $-2$ as well.
For training DQN, we collect $2000$ training sequences in each iteration.
In this experiment, we also compare with the DQN using MLP as its Q-network.
For the MLP scheme, we consider two MLP structures.
To be specific, DQN+MLP1 has a MLP with one hidden layer, and DQN+MLP2 has a MLP with two hidden layers.
All the hidden layers contain $64$ neurons, and the number of neurons of RC is also set to $64$.
The learning rates of DQN+RC, DQN+MLP1, DQN+MLP2, and Q-learning are set to $0.01$.
We can observe from Fig.~\ref{fig:result_channel6_su2_punish_-2} that the myopic method has better performance than all learning-based approaches at the beginning of training because it has full knowledge of the system dynamics.
However, the myopic method can not benefit the SU from learning the access strategy of other SUs. 
Therefore, the myopic method will result a constant collision rate with other SUs.
On the other hand, the learning-based methods are able to gradually learn from the stochastic environment and the interactions among SUs and PUs, therefore, learning-based methods all develop strategies that have higher average reward than the myopic method. 
As shown in Fig.\ref{fig:result_channel6_su2_punish_-2}, all learning-based methods have $0$ collision among SUs when their training curves converge, meaning each SU learns the access strategy of other SUs perfectly. 
The performance comparison among learning-based methods reflects the following relationship: $\text{DQN+RC} \approx \text{Q-learning} > \text{DQN+MLP1} \approx \text{DQN+MLP2}$.
To be specific, Q-learning and DQN+RC have similar performance because the number of channels is only $6$ in this experiment.
Clearly, DQN+RC outperforms both DQN+MLP1 and DQN+MLP2.
This is because the recurrent structure of RC enables a more complete construction of a high dimensional functional space to approximate the network state space than MLP, even though both RC and MLP have an equal number of neurons.
An interesting observation is that DQN+MLP2 performs slightly worse than DQN+MLP1 even DQN+MLP2 has more complicated structure.
This is mainly because a deeper network usually needs more training data and training time to make it converge. 
Therefore, having more neurons in a neural network does not always translate to having better performance.

\subsection{Comparison Between DQN+RC and DQN+MLP1}
In this section, we demonstrate the impact of temporal correlation of the spectrum sensing outcomes on the learning performance of DQN+MLP1 and DQN+RC. 
Note that we do not include DQN+MLP2 in this comparison since the performance of DQN+MLP1 is similar to DQN+MLP2 as discussed in Section 5.2.
It is important to note that in reality the spectrum sensing outcomes are correlated over time due to 1) PUs' activities have temporal correlation and 2) the wireless channel demonstrates temporal correlation. 
A simple experiment is designed where the PUs' activities is time dependent.
In the experiment, we assume that there is only one channel, one PU, and one SU for the underlying DSA network.
The number of training sequences in each iteration is set to $2000$.
Instead of using the memoryless two-state Markov chain to model the PU's behavior, we assume PU's activity is time-dependent (that is, it is governed by a process which has memory). 
For simplicity, we assume that a PU will access the channel once every three time slots with $\{\textit{Inactive}, \textit{Inactive}, \textit{Active}\}$ as a period.
\begin{figure}[ht]
\centering
	\includegraphics[width=.95\linewidth]{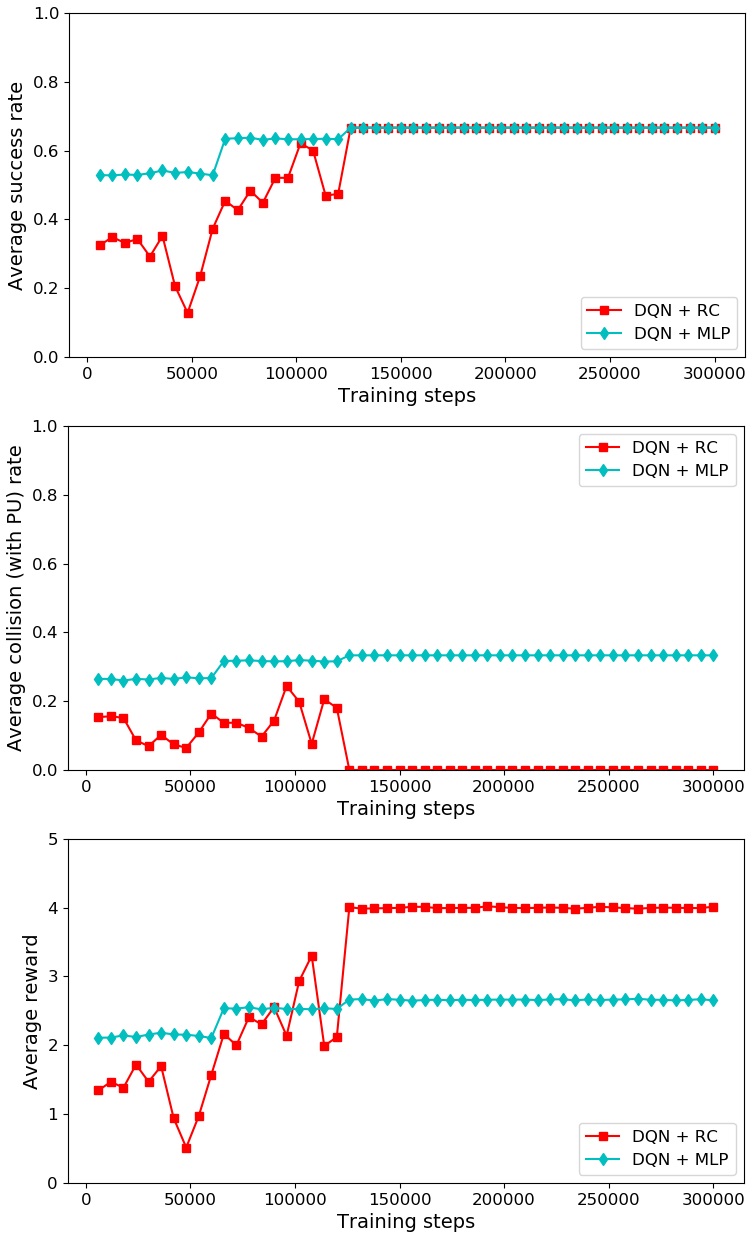}
	\caption{The average success number, collision number with PUs, and reward with the training time (No. of channel = 1 and No. of SU = 1)}
	\label{fig:result_channel_1_su_1}
\end{figure}
The structure of the MLP1 contains one hidden layer of $64$ neurons. 
To make a fair comparison, the number of neurons of RC is also set to $64$. Besides, the learning rate of both MLP1 and RC is $0.01$. 
The results are shown in Fig.~\ref{fig:result_channel_1_su_1}.
Because the expected reward of access the channel is higher than not accessing the channel, the strategy of DQN+MLP1 is to always access the channel regardless of spectrum sensing results, resulting in the success rate to be $1/3$ and the collision rate to be $2/3$.
On the other hand, DQN+RC could precisely learn the activity of the PU to access the channel only when the PU is \textit{Inactive}.
This is because DQN+RC can effectively learn the temporal correlation of the sensing outcome, which can be used to improve access performance.
The experiment result clearly demonstrates that RC is able to learn the underlying temporal correlation, while MLP can only learn the one-to-one mapping from state to action.

\section{Conclusions}
In this paper, we study spectrum access strategies in a distributive DSA network under the condition of imperfect spectrum sensing and no centralized controllers. 
A new DSA strategy based on the combination of DQN and RC is introduced, in which DRL is employed to learn the availability of spectrum resources, and RC is utilized to realize DQN by taking advantage of the underlying temporal correlation of the DSA network. 
Equipped with our DSA strategy, each SU is able to make proper spectrum access decisions distributedly relying only on minimal broadcast information from the PUs, their own spectrum sensing outcomes, as well as the learning outcomes. 
Experiments are designed to verify the performance of our proposed strategy. 
The experiment results demonstrate that our learning method could provide higher successful transmission rate and lower transmission collision rate as oppose to the myopic strategy which generally assumes channel statistics.
As compared to the Q-learning, our DQN based approach is shown to converge faster with better performance. 
Due to the recurrent structure of the RC, our DQN+RC-based scheme can outperform DQN+MLP-based scheme when spectrum sensing outcomes exhibit temporal correlation.

\section*{Acknowledgement}
This work was supported in part by National Science Foundation under grant ECCS-1731672 and ECCS-1811497.

\ifCLASSOPTIONcaptionsoff
  \newpage
\fi

\bibliography{eqbib}

\begin{thebibliography}{10}
\providecommand{\url}[1]{#1}
\csname url@samestyle\endcsname
\providecommand{\newblock}{\relax}
\providecommand{\bibinfo}[2]{#2}
\providecommand{\BIBentrySTDinterwordspacing}{\spaceskip=0pt\relax}
\providecommand{\BIBentryALTinterwordstretchfactor}{4}
\providecommand{\BIBentryALTinterwordspacing}{\spaceskip=\fontdimen2\font plus
\BIBentryALTinterwordstretchfactor\fontdimen3\font minus
  \fontdimen4\font\relax}
\providecommand{\BIBforeignlanguage}[2]{{%
\expandafter\ifx\csname l@#1\endcsname\relax
\typeout{** WARNING: IEEEtran.bst: No hyphenation pattern has been}%
\typeout{** loaded for the language `#1'. Using the pattern for}%
\typeout{** the default language instead.}%
\else
\language=\csname l@#1\endcsname
\fi
#2}}
\providecommand{\BIBdecl}{\relax}
\BIBdecl

\bibitem{Cisco}
M.~McHenry, ``Cisco visual networking index: Forecast and methodology,
  2016–2021,'' Cisco Whitepaper, Tech. Rep., Jun. 2003.

\bibitem{yin2012mining}
S.~Yin, D.~Chen, Q.~Zhang, M.~Liu, and S.~Li, ``Mining spectrum usage data: a
  large-scale spectrum measurement study,'' \emph{{IEEE} Trans. Mobile
  Comput.}, vol.~11, no.~6, pp. 1033--1046, Jun. 2012.

\bibitem{Islam2008spectrum}
M.~H. Islam, C.~L. Koh, S.~W. Oh, X.~Qing, Y.~Y. Lai, C.~Wang, Y.~Liang, B.~E.
  Toh, F.~Chin, G.~L. Tan, and W.~Toh, ``Spectrum survey in singapore:
  Occupancy measurements and analyses,'' in \emph{2008 3rd International
  Conference on Cognitive Radio Oriented Wireless Networks and Communications
  (CrownCom 2008)}, May 2008, pp. 1--7.

\bibitem{mchenry2006chicago}
M.~A. McHenry, P.~A. Tenhula, D.~McCloskey, D.~A. Roberson, and C.~S. Hood,
  ``Chicago spectrum occupancy measurements \& analysis and a long-term studies
  proposal,'' in \emph{Proceedings of the first international workshop on
  TAPAS}, no.~1.\hskip 1em plus 0.5em minus 0.4em\relax ACM, Aug. 2006.

\bibitem{SSC}
``\url{http://www.sharedspectrum.com}.''\hskip 1em plus 0.5em minus 0.4em\relax
  Shared Spectrum Company.

\bibitem{force2002spectrum}
P.~J. Kolodzy, ``Spectrum policy task force report,'' FCC, Tech. Rep., Nov.
  2002.

\bibitem{pi2011introduction}
Z.~Pi and F.~Khan, ``An introduction to millimeter-wave mobile broadband
  systems,'' \emph{{IEEE} Commun. Mag.}, vol.~49, no.~6, pp. 101--107, Jun.
  2011.

\bibitem{FCC}
``\url{http://transition.fcc.gov/Daily_Releases/Daily_Business/2016/db0714/DOC-340310A1.pdf}.''\hskip
  1em plus 0.5em minus 0.4em\relax FCC.

\bibitem{bhattarai2016overview}
S.~Bhattarai, J.-M.~J. Park, B.~Gao, K.~Bian, and W.~Lehr, ``An overview of
  dynamic spectrum sharing: Ongoing initiatives, challenges, and a roadmap for
  future research,'' \emph{{IEEE} Trans. on Cogn. Commun. Netw.}, vol.~2,
  no.~2, pp. 110--128, Jun. 2016.

\bibitem{federal2015report}
``Report and order and second further notice of proposed rulemaking,'' FCC,
  Tech. Rep., 2015.

\bibitem{liao2014listen}
Y.~Liao, T.~Wang, L.~Song, and Z.~Han, ``Listen-and-talk: Full-duplex cognitive
  radio networks,'' in \emph{GLOBECOM}, Dec. 2014, pp. 3068--3073.

\bibitem{fu2017throughput}
Z.~Fu, W.~Xu, Z.~Feng, X.~Lin, and J.~Lin, ``Throughput analysis of
  lte-licensed-assisted access networks with imperfect spectrum sensing,'' in
  \emph{WCNC}, May 2017, pp. 1--6.

\bibitem{stotas2011enhancing}
S.~Stotas and A.~Nallanathan, ``Enhancing the capacity of spectrum sharing
  cognitive radio networks,'' \emph{{IEEE} Trans. Veh. Technol.}, vol.~60,
  no.~8, pp. 3768--3779, Oct. 2011.

\bibitem{Li2010Qlearning}
H.~Li, ``Multi-agent q-learning for competitive spectrum access in cognitive
  radio systems,'' in \emph{Fifth IEEE Workshop on Networking Technologies for
  Software Defined Radio Networks}, Jun. 2010, pp. 1--6.

\bibitem{Wang2018DRL}
S.~Wang, H.~Liu, P.~H. Gomes, and B.~Krishnamachari, ``Deep reinforcement
  learning for dynamic multichannel access in wireless networks,'' \emph{{IEEE}
  Trans. on Cogn. Commun. Netw.}, vol.~4, no.~2, pp. 257--265, Jun. 2018.

\bibitem{Naparstek2017DRL}
O.~Naparstek and K.~Cohen, ``Deep multi-user reinforcement learning for dynamic
  spectrum access in multichannel wireless networks,'' in \emph{GLOBECOM}, Dec.
  2017, pp. 1--7.

\bibitem{2013RNNdifficulty}
R.~Pascanu, T.~Mikolov, and Y.~Bengio, ``On the difficulty of training
  recurrent neural networks,'' in \emph{ICML}, 2013, pp. 1310--1318.

\bibitem{2009RCtoRNN}
M.~Luko{\v{s}}evi{\v{c}}ius and H.~Jaeger, ``Reservoir computing approaches to
  recurrent neural network training,'' \emph{Computer Science Review}, vol.~3,
  no.~3, pp. 127--149, 2009.

\bibitem{sak2014long}
H.~Sak, A.~Senior, and F.~Beaufays, ``Long short-term memory recurrent neural
  network architectures for large scale acoustic modeling,'' in \emph{ISCA},
  Sep. 2014, pp. 338--342.

\bibitem{MIMOOFDM2017}
S.~Mosleh, L.~Liu, C.~Sahin, Y.~R. Zheng, and Y.~Yi, ``Brain-inspired wireless
  communications: Where reservoir computing meets mimo-ofdm,'' \emph{{IEEE}
  Trans. Neural Netw. Learn. Syst.}, no.~99, pp. 1--15, Dec. 2017.

\bibitem{mnih2015DRL}
V.~Mnih, K.~Kavukcuoglu, D.~Silver, A.~A. Rusu, J.~Veness, M.~G. Bellemare,
  A.~Graves, M.~Riedmiller, A.~K. Fidjeland, G.~Ostrovski, S.~Petersen,
  C.~Beattie, A.~Sadik, I.~Antonoglou, H.~King, D.~Kumaran, D.~Wierstra,
  S.~Legg, and D.~Hassabis, ``Human-level control through deep reinforcement
  learning,'' \emph{Nature}, vol. 518, no. 7540, pp. 529--533, Feb. 2015.

\bibitem{jaeger2001echo}
H.~Jaeger, ``The “echo state” approach to analysing and training recurrent
  neural networks-with an erratum note,'' \emph{German National Research Center
  for Information Technology GMD Technical Report}, vol. 148, no.~34, p.~13,
  Jan. 2001.

\bibitem{1992QLearning}
C.~J. C.~H. Watkins and P.~Dayan, ``Q-learning,'' \emph{Machine Learning},
  vol.~8, no.~3, pp. 279--292, May. 1992.

\bibitem{egreedy}
E.~Rodrigues~Gomes and R.~Kowalczyk, ``Dynamic analysis of multiagent
  q-learning with $\varepsilon$-greedy exploration,'' in \emph{ICML}, Jun.
  2009, pp. 369--376.

\bibitem{2013atari}
V.~Mnih, K.~Kavukcuoglu, D.~Silver, A.~Graves, I.~Antonoglou, D.~Wierstra, and
  M.~Riedmiller, ``Playing atari with deep reinforcement learning,''
  \emph{arXiv preprint arXiv:1312.5602}, 2013.

\bibitem{mikolov2010recurrent}
T.~Mikolov, M.~Karafi{\'a}t, L.~Burget, J.~{\v{C}}ernock{\`y}, and
  S.~Khudanpur, ``Recurrent neural network based language model,'' in
  \emph{ISCA}, Sep. 2010, pp. 1045--1048.

\bibitem{bengio1994learning}
Y.~Bengio, P.~Simard, and P.~Frasconi, ``Learning long-term dependencies with
  gradient descent is difficult,'' \emph{{IEEE} Trans. Neural Netw.}, vol.~5,
  no.~2, pp. 157--166, Mar. 1994.

\bibitem{jaeger2002tutorial}
H.~Jaeger, \emph{Tutorial on training recurrent neural networks, covering BPPT,
  RTRL, EKF and the" echo state network" approach}.\hskip 1em plus 0.5em minus
  0.4em\relax GMD-Forschungszentrum Informationstechnik Bonn, 2002, vol.~5.

\bibitem{enel2016reservoir}
P.~Enel, E.~Procyk, R.~Quilodran, and P.~F. Dominey, ``Reservoir computing
  properties of neural dynamics in prefrontal cortex,'' \emph{PLoS
  computational biology}, vol.~12, no.~6, p. e1004967, Jun. 2016.

\bibitem{jaeger2007echo}
H.~Jaeger, ``Echo state network,'' \emph{Scholarpedia}, vol.~2, no.~9, p. 2330,
  2007.

\bibitem{kian2018DFR}
K.~Hamedani, L.~Liu, R.~Atat, J.~Wu, and Y.~Yi, ``Reservoir computing meets
  smart grids: Attack detection using delayed feedback networks,'' \emph{{IEEE}
  Trans. Ind. Informat.}, vol.~14, no.~2, pp. 734--743, Feb. 2018.

\bibitem{Mingzhe2017ESN}
M.~Chen, W.~Saad, C.~Yin, and M.~Debbah, ``Echo state networks for proactive
  caching in cloud-based radio access networks with mobile users,''
  \emph{{IEEE} Trans. Wireless Commun.}, vol.~16, no.~6, pp. 3520--3535, Jun.
  2017.

\bibitem{meinila2009winner}
J.~Meinil{\"a}, P.~Ky{\"o}sti, T.~J{\"a}ms{\"a}, and L.~Hentil{\"a}, ``Winner
  {II} channel models,'' \emph{Radio Technologies and Concepts for
  IMT-Advanced}, pp. 39--92, 2009.

\bibitem{Liu2010Myopic}
K.~Liu, Q.~Zhao, and B.~Krishnamachari, ``Dynamic multichannel access with
  imperfect channel state detection,'' \emph{{IEEE} Trans. Signal Process.},
  vol.~58, no.~5, pp. 2795--2808, May 2010.

\bibitem{Li2012Myopic}
Y.~Li, S.~K. Jayaweera, M.~Bkassiny, and K.~A. Avery, ``Optimal myopic sensing
  and dynamic spectrum access in cognitive radio networks with low-complexity
  implementations,'' \emph{{IEEE} Trans. Wireless Commun.}, vol.~11, no.~7, pp.
  2412--2423, Jul. 2012.

\bibitem{2010RL}
K.~L.~A. Yau, P.~Komisarczuk, and P.~D. Teal, ``Applications of reinforcement
  learning to cognitive radio networks,'' in \emph{ICC Workshops}, May 2010,
  pp. 1--6.

\bibitem{Morozs2016Qlearning}
N.~Morozs, T.~Clarke, and D.~Grace, ``Distributed heuristically accelerated
  q-learning for robust cognitive spectrum management in lte cellular
  systems,'' \emph{{IEEE} Trans. Mobile Comput.}, vol.~15, no.~4, pp. 817--825,
  Apr. 2016.

\end{thebibliography}
\bibliographystyle{IEEEtran}

\begin{IEEEbiography}
    [{\includegraphics[width=1in,height=1.25in,clip,keepaspectratio]{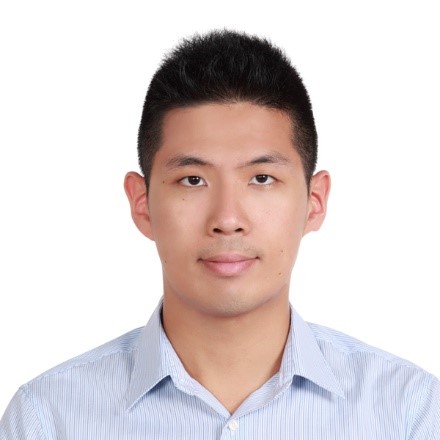}}]{Hao-Hsuan Chang}
received the B.Sc. degree in electrical engineering and the M.S. degree in communication engineering from National Taiwan University, Taipei, Taiwan in 2014 and 2015. He is currently pursuing the Ph.D. degree in electrical and computer engineering at Virginia Polytechnic Institute and State University, Blacksburg, VA, USA. His research interests include dynamic spectrum access, deep learning, and machine learning.
\end{IEEEbiography}

\begin{IEEEbiography}
    [{\includegraphics[width=1in,height=1.25in,clip,keepaspectratio]{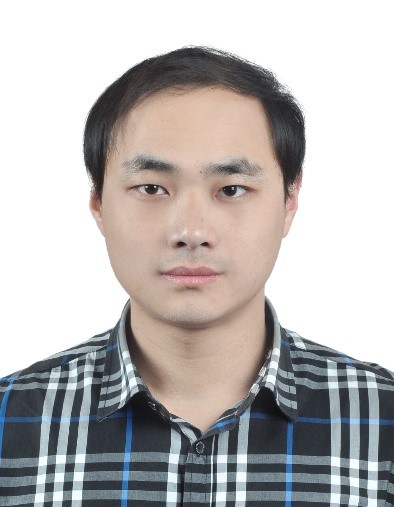}}]{Hao Song}
(S’14) received the B.E. degree in electrical information engineering from Zhengzhou University, Zhengzhou, China, in 2011. He is currently pursuing the Ph.D. degree with the Bradley Department of Electrical and Computer Engineering, Virginia Tech, Blacksburg, VA, USA. His research interests focus on mobile network optimization, dynamic spectrum access (DSA), and machine learning and its applications in wireless communications.
\end{IEEEbiography}

\begin{IEEEbiography}
    [{\includegraphics[width=1in,height=1.25in,clip,keepaspectratio]{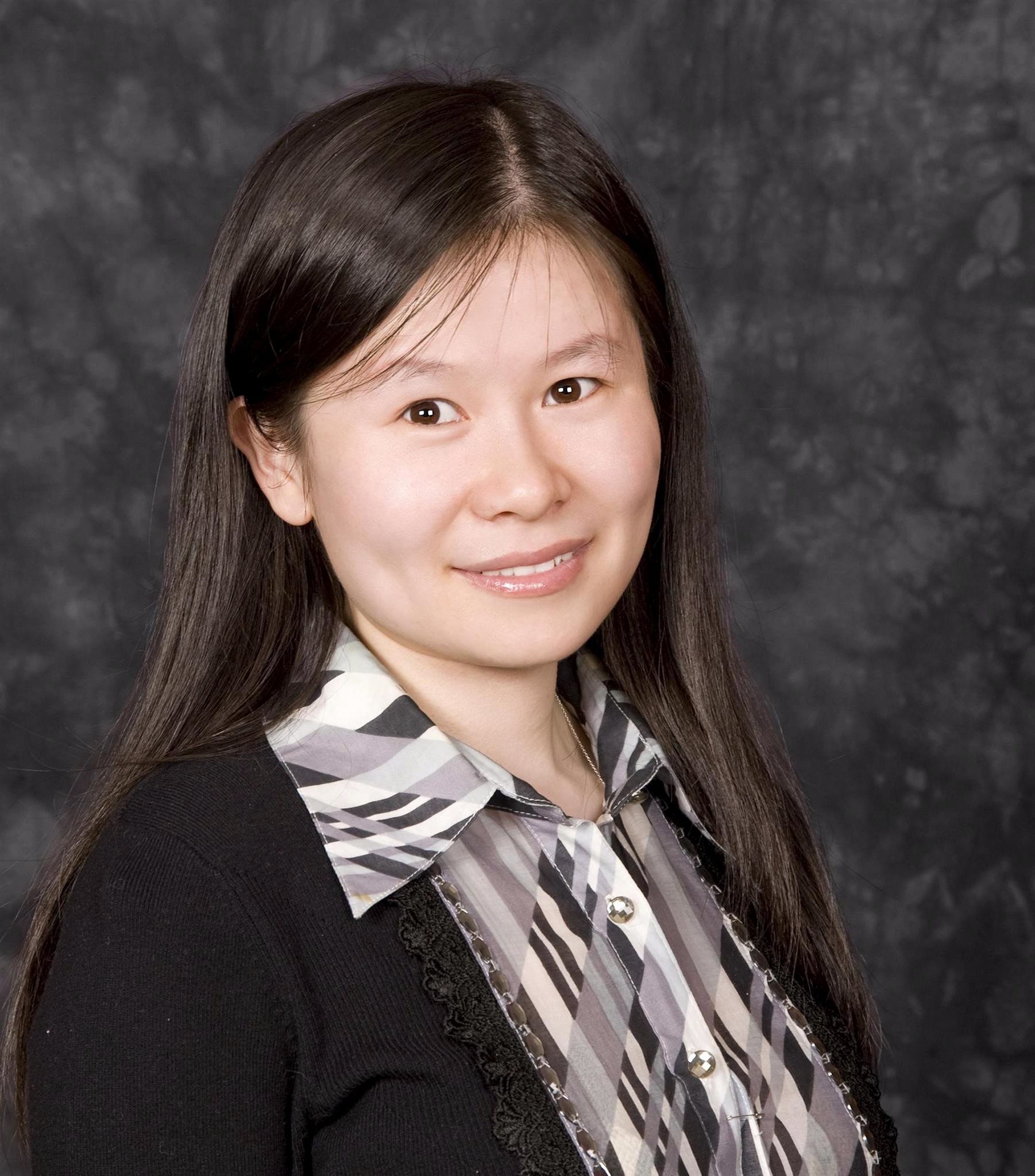}}]{Yang Yi}
(SM’17) is an Assistant Professor in the Bradley Department of ECE at Virginia Tech (VT). She received the B.S. and M.S. degrees in electronic engineering at Shanghai Jiao Tong University, and the Ph.D. degree in electrical and computer engineering at Texas A\&M University. Her research interests include very large scale integrated (VLSI) circuits and systems, computer aided design (CAD), and neuromorphic computing. Dr. Yi is currently serving as an associate editor for cyber journal of selected areas in microelectronics and has been serving on the editorial board of international journal of computational \& neural engineering. Dr. Yi is the recipient of 2018 National Science CAREER award, 2016 Miller Professional Development Award for Distinguished Research, 2016 United States Air Force (USAF) Summer Faculty Fellowship, 2015 NSF EPSCoR First Award, and 2015 Miller Scholar.
\end{IEEEbiography}

\begin{IEEEbiography}
    [{\includegraphics[width=1in,height=1.25in,clip,keepaspectratio]{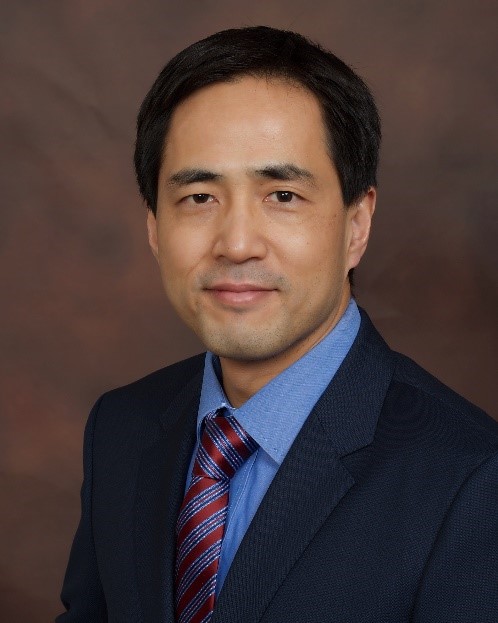}}]{Jiazhong (Charlie) Zhang}
(F’16) is a VP and head of the Standards and Mobility Innovation Lab of Samsung Research America, where he leads research, prototyping, and standards for 5G and future multimedia networks. From 2009 to 2013, he served as the Vice Chairman of the 3GPP RAN1 working group and led development of LTE and LTE-Advanced technologies such as 3D channel modeling, UL-MIMO, CoMP, Carrier Aggregation for TD-LTE. He received his Ph.D. degree from the University of Wisconsin, Madison.
\end{IEEEbiography}

\begin{IEEEbiography}
    [{\includegraphics[width=1in,height=1.25in,clip,keepaspectratio]{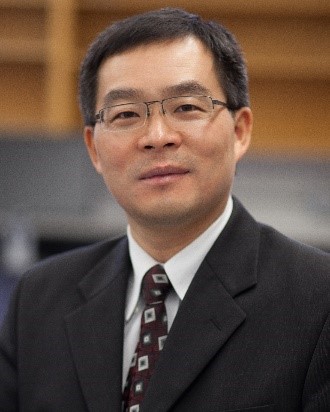}}]{Haibo He}
(SM'11-F’18) received the B.S. and M.S. degrees in electrical engineering from Huazhong University of Science and Technology, China, in 1999 and 2002, respectively, and the Ph.D. degree in electrical engineering from Ohio University in 2006. From 2006 to 2009, he was an Assistant Professor at the Department of Electrical and Computer Engineering at Stevens Institute of Technology. Currently, he is the Robert Haas Endowed Chair Professor at the Department of Electrical, Computer, and Biomedical Engineering at the University of Rhode Island. 

His research interests include adaptive learning and control, computational intelligence, machine learning, data mining, and various applications. He has published 1 sole-author research book (Wiley), edited 1 book (Wiley-IEEE) and 6 conference proceedings (Springer), and authored and co-authored over 300 peer-reviewed journal and conference papers. He has delivered more than 100 invited talks around the globe. He was the Chair of IEEE Computational Intelligence Society (CIS) Emergent Technologies Technical Committee (ETTC) (2015) and the Chair of IEEE CIS Neural Networks Technical Committee (NNTC) (2013 and 2014). He served as the General Chair of 2014 IEEE Symposium Series on Computational Intelligence (IEEE SSCI’14, Orlando, Florida). He was a recipient of the IEEE International Conference on Communications Best Paper Award (2014), IEEE Computational Intelligence Society (CIS) Outstanding Early Career Award (2014), and National Science Foundation (NSF) CAREER Award (2011). Currently, he is the Editor-in-Chief of the IEEE Transactions on Neural Networks and Learning Systems.
\end{IEEEbiography}

\begin{IEEEbiography}
    [{\includegraphics[width=1in,height=1.25in,clip,keepaspectratio]{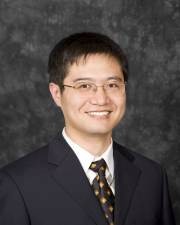}}]{Lingjia Liu}
(SM’15) received his B.S. degree in Electronic Engineering from Shanghai Jiao Tong University and Ph.D. degree in Electrical and Computer Engineering from Texas A\&M University. Prior to joining the ECE Department at Virginia Tech (VT), he was an Associate Professor in the EECS Department at the University of Kansas (KU). He spent 4+ years working in Mitsubishi Electric Research Laboratory (MERL) and the Standards \& Mobility Innovation Lab of Samsung Research America (SRA) where he received Global Samsung Best Paper Award in 2008 and 2010. He was leading Samsung’s efforts on multiuser MIMO, CoMP, and HetNets in LTE/LTE-Advanced standards. 

Dr. Liu’s general research interests mainly lie in emerging technologies for 5G cellular networks including machine learning for wireless networks, massive MIMO, massive MTC communications, and mmWave communications. Dr. Liu received Air Force Summer Faculty Fellow from 2013 to 2017, Miller Scholar at KU in 2014, Miller Professional Development Award for Distinguished Research at KU in 2015, 2016 IEEE GLOBECOM Best Paper Award, 2018 IEEE ISQED Best Paper Award, and 2018 IEEE TAOS Best Paper Award.
\end{IEEEbiography}

\end{document}